\documentclass[conference]{IEEEtran}
\IEEEoverridecommandlockouts
\usepackage{cite}
\usepackage{amsmath,amssymb,amsfonts}
\usepackage{algorithmic}
\usepackage{graphicx}
\usepackage{tikz}
\usetikzlibrary{colorbrewer}
\usepackage{textcomp}
\usepackage{xcolor}
\usepackage{subcaption}
\usepackage{siunitx}
\usepackage{cite}
\usepackage{pifont}% http://ctan.org/pkg/pifont
\newcommand{\cmark}{\ding{51}}%

\newcommand{\xmark}{\ding{55}}%
\usepackage{multirow}
\usepackage{diagbox}
\usepackage{makecell}
\usepackage{booktabs}
\usepackage{array}
\usepackage{scalerel}
\usepackage{tikz}
\usetikzlibrary{svg.path}
\usepackage{pgfplots}
\pgfplotsset{select coords between index/.style 2 args={
		x filter/.code={
			\ifnum\coordindex<#1\fi
			\ifnum\coordindex>#2\fi
		}
}}
\definecolor{bargrey}{RGB}{144,144,144}
\definecolor{barpink}{RGB}{188,80,144}
\definecolor{barred}{RGB}{255,99,97}
\definecolor{barorange}{RGB}{255,166,0}
\definecolor{barblue}{RGB}{88,80,141}
\definecolor{barblue2}{RGB}{0,63,92}
\usepgfplotslibrary{groupplots}

\newcolumntype{P}[1]{>{\centering\arraybackslash}m{#1}}
\newcolumntype{H}[1]{>{\setbox0=\hbox\bgroup}c<{\egroup}@{}} % hide column

\def\BibTeX{{\rm B\kern-.05em{\sc i\kern-.025em b}\kern-.08em
    T\kern-.1667em\lower.7ex\hbox{E}\kern-.125emX}}
\def\camera#1#2#3{
  \begin{scope}[shift={#1}, rotate=#2, scale=#3, every node/.append style={transform shape}]
    \draw [fill=Paired-I, draw=Paired-J, thick](0,0) -- (2,2.5) -- (-2,2.5) -- cycle;
    \draw [fill=white, draw=Paired-J, thick](0,0) circle (1);
  \end{scope}
}

\usepackage{hyperref}
\hypersetup{
	colorlinks=true,	
	pdfstartview=FitV,	
	linkcolor=black!50!blue,	
	citecolor=black!50!green,	
	urlcolor=black,		
	breaklinks=true, 	
}

\definecolor{orcidlogocol}{HTML}{A6CE39}
\tikzset{
  orcidlogo/.pic={
    \fill[orcidlogocol] svg{M256,128c0,70.7-57.3,128-128,128C57.3,256,0,198.7,0,128C0,57.3,57.3,0,128,0C198.7,0,256,57.3,256,128z};
    \fill[white] svg{M86.3,186.2H70.9V79.1h15.4v48.4V186.2z}
                 svg{M108.9,79.1h41.6c39.6,0,57,28.3,57,53.6c0,27.5-21.5,53.6-56.8,53.6h-41.8V79.1z M124.3,172.4h24.5c34.9,0,42.9-26.5,42.9-39.7c0-21.5-13.7-39.7-43.7-39.7h-23.7V172.4z}
                 svg{M88.7,56.8c0,5.5-4.5,10.1-10.1,10.1c-5.6,0-10.1-4.6-10.1-10.1c0-5.6,4.5-10.1,10.1-10.1C84.2,46.7,88.7,51.3,88.7,56.8z};
  }
}

\newcommand\orcidicon[1]{\href{https://orcid.org/#1}{\mbox{\scalerel*{
\begin{tikzpicture}[yscale=-1,transform shape]
\pic{orcidlogo};
\end{tikzpicture}
}{|}}}}

\begin{document}

\title{RadarScenes: A Real-World Radar Point Cloud Data Set for Automotive Applications}

\author{\IEEEauthorblockN{Ole Schumann\IEEEauthorrefmark{1}\orcidicon{0000-0001-8953-1075}, Markus Hahn\IEEEauthorrefmark{4}, Nicolas Scheiner\IEEEauthorrefmark{1}\orcidicon{0000-0002-5176-6159}, Fabio Weishaupt\IEEEauthorrefmark{1}\orcidicon{0000-0001-6489-0522},\\
Julius F. Tilly\IEEEauthorrefmark{1}, J\"urgen Dickmann\IEEEauthorrefmark{1}\orcidicon{0000-0002-4328-3368}, and Christian W\"ohler\IEEEauthorrefmark{3}\orcidicon{0000-0002-0715-0955}}
\IEEEauthorblockA{\IEEEauthorrefmark{1}\textit{Environment Perception, Mercedes-Benz AG}, Stuttgart, Germany, Email: \href{mailto:ole.schumann@mercedes-benz.com}{ole.schumann@mercedes-benz.com}}
\IEEEauthorblockA{\IEEEauthorrefmark{4}\textit{Continental AG, during his time at Daimler AG}, Ulm, Germany}
\IEEEauthorblockA{\IEEEauthorrefmark{3}\textit{Faculty of Electrical Engineering \& Information Technology, TU Dortmund}, Dortmund, Germany}}

\maketitle

\begin{abstract}
A new automotive radar data set with measurements and point-wise annotations from more than four hours of driving is presented.
Data provided by four series radar sensors mounted on one test vehicle were
recorded and the individual detections of dynamic objects were manually grouped
to clusters and labeled afterwards.
The purpose of this data set is to enable the development of novel (machine learning-based) radar perception algorithms with the focus on moving road users.
Images of the recorded sequences were captured using a documentary camera.
For the evaluation of future object detection and classification algorithms, proposals for score calculation are made so that researchers can evaluate their algorithms on a common basis.
Additional information as well as download instructions can be found on the website of the data set: \url{www.radar-scenes.com}.
\end{abstract}

\begin{IEEEkeywords}
dataset, radar, machine learning, classification %, data
\end{IEEEkeywords}

\section{Introduction} \label{sec:intro}
On the way to autonomous driving, a highly accurate and reliable perception of
the vehicle's environment is essential.
Different sensor modalities are typically combined to complement each other.
This leads to a presence of radar sensors in every major self-driving vehicle setup.
Due to its operation in the millimeter wave frequency band, radar is known for
its robustness in adverse weather conditions. Furthermore, the capability of
directly measuring relative velocity via evaluation of the Doppler effect is
unique in today's sensor stacks. 
Both taken together and combined with moderate costs, radar sensors have been integrated in a wide variety of series production cars for several years now.

\begin{figure}[t!]
	\centering
	\begin{subfigure}{.32\linewidth}
		\includegraphics[width=\textwidth]{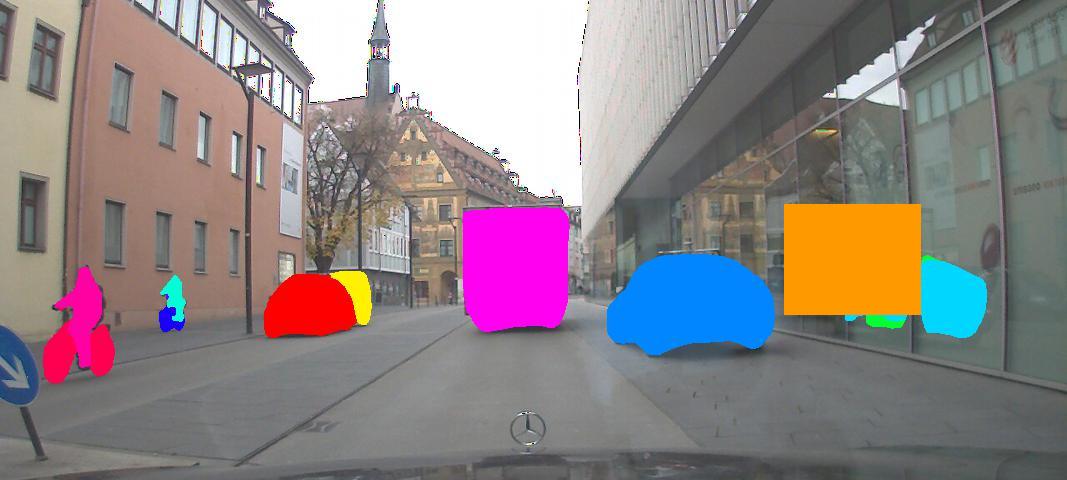}
	\end{subfigure}
	\begin{subfigure}{.32\linewidth}
	\includegraphics[width=\textwidth]{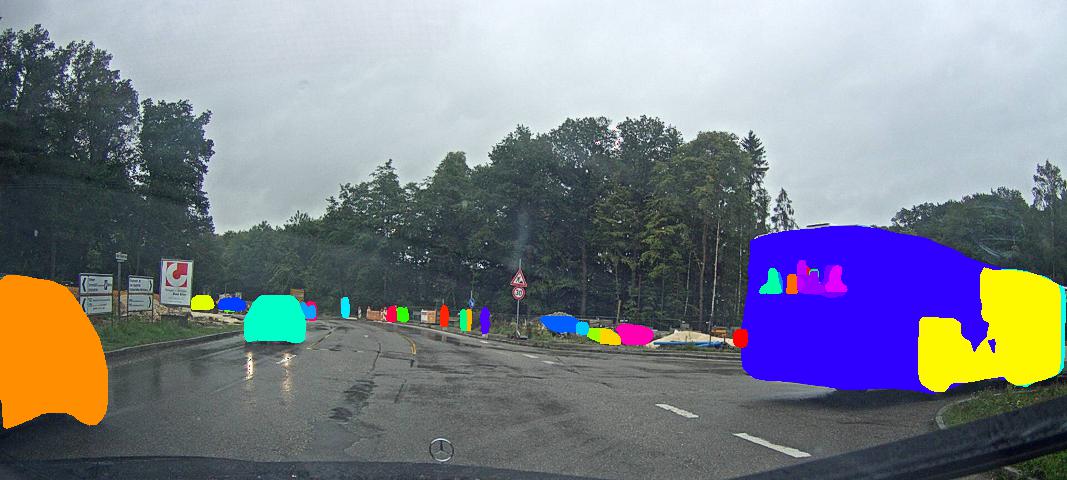}
	\end{subfigure}
	\begin{subfigure}{.32\linewidth}
	\includegraphics[width=\textwidth]{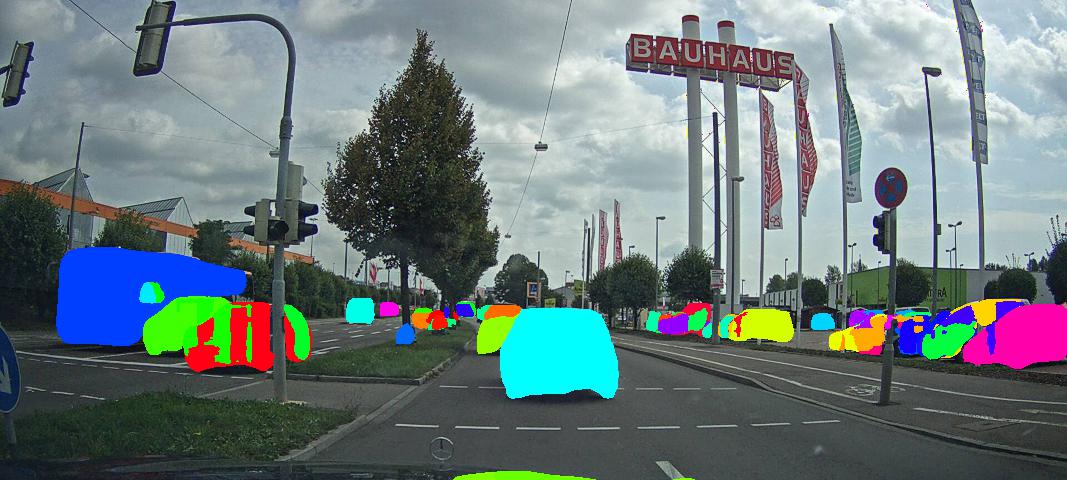}
	\end{subfigure}
	\vspace{.5em}
	
	\begin{subfigure}{.32\linewidth}
		\frame{\includegraphics[width=\textwidth]{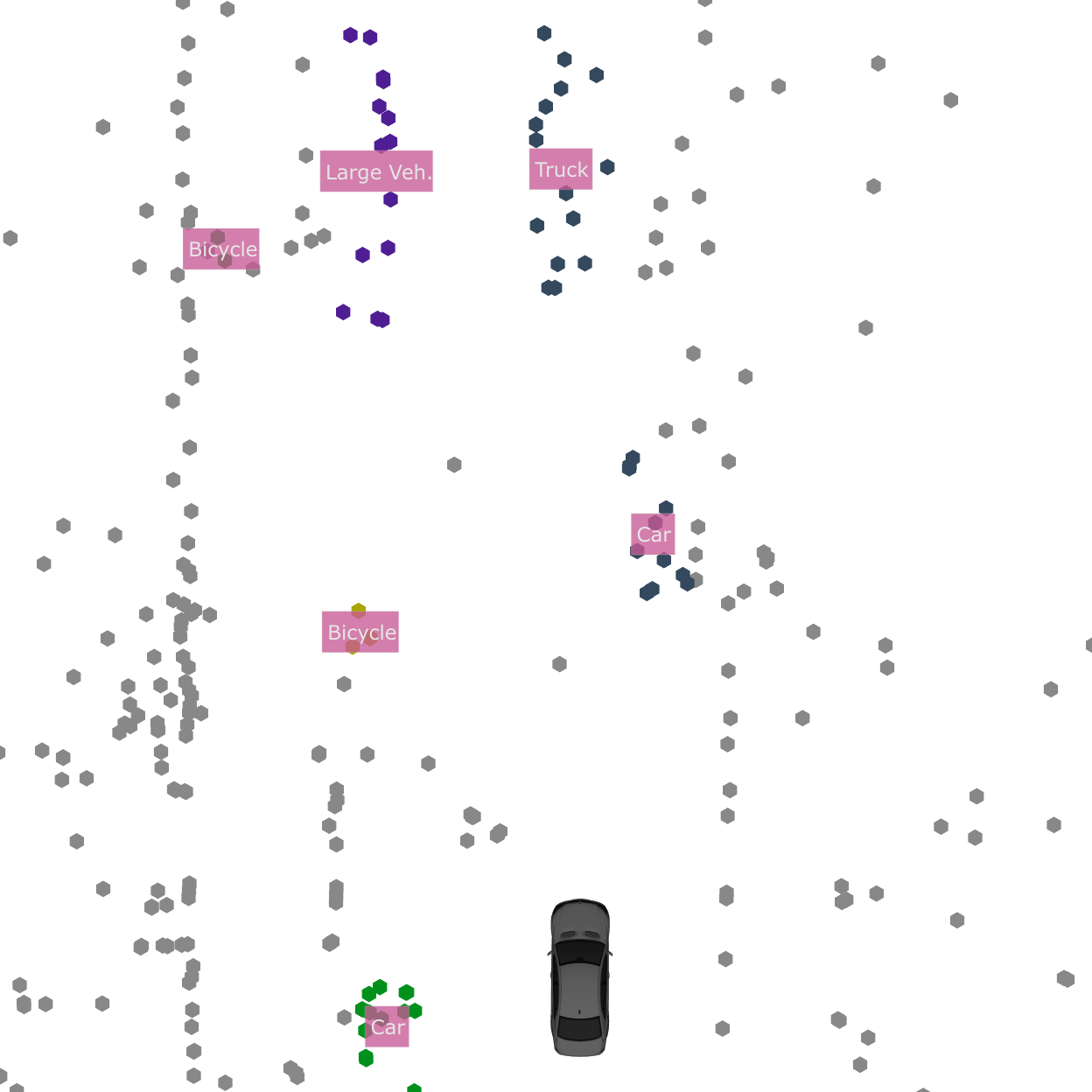}}
		\captionsetup{skip=1pt, font=footnotesize} \caption*{{Inner City}}
	\end{subfigure}
	\begin{subfigure}{.32\linewidth}
	\frame{\includegraphics[width=\textwidth]{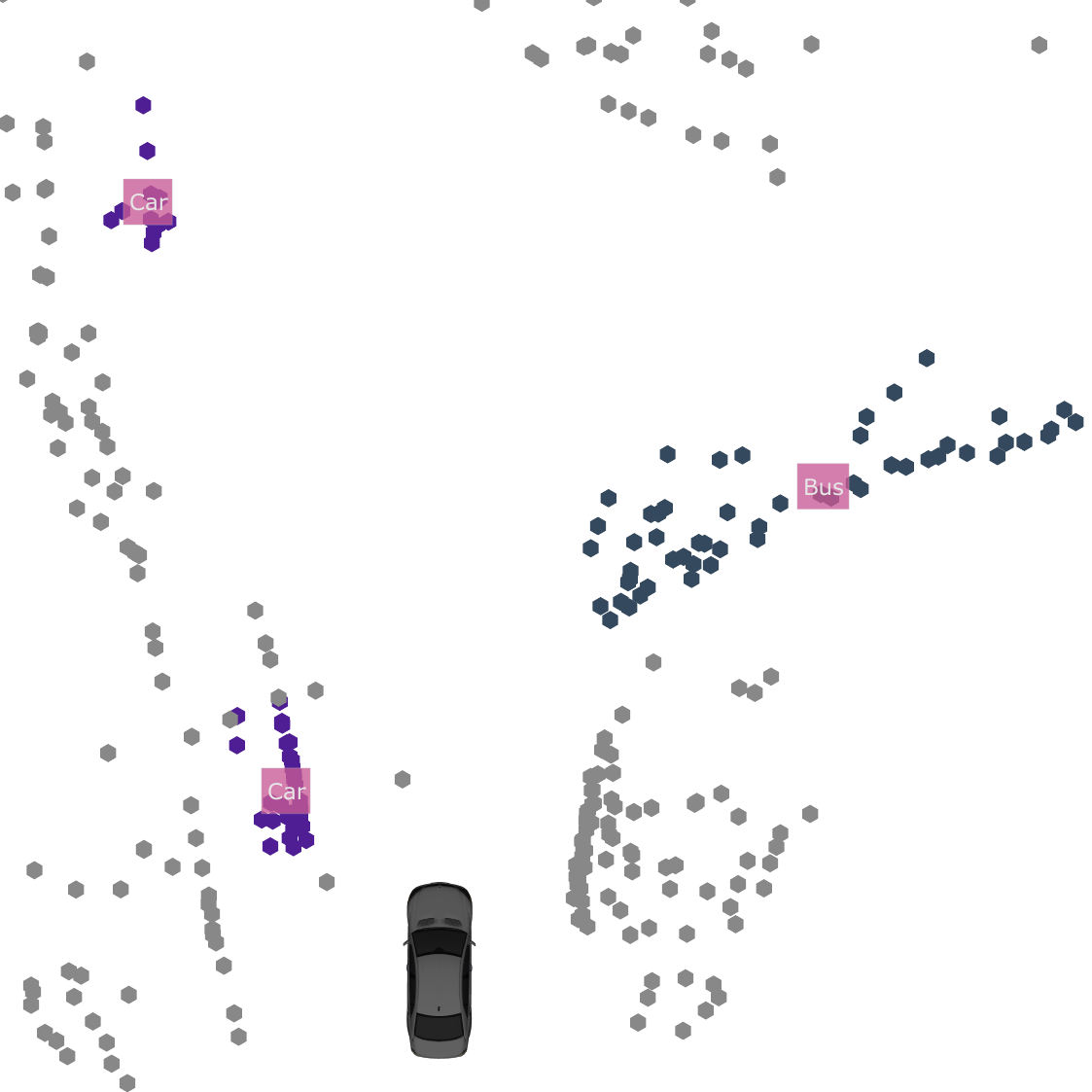}}
	\captionsetup{skip=1pt, font=footnotesize} \caption*{{T-junction}}
	\end{subfigure}
		\begin{subfigure}{.32\linewidth}
		\frame{\includegraphics[width=\textwidth]{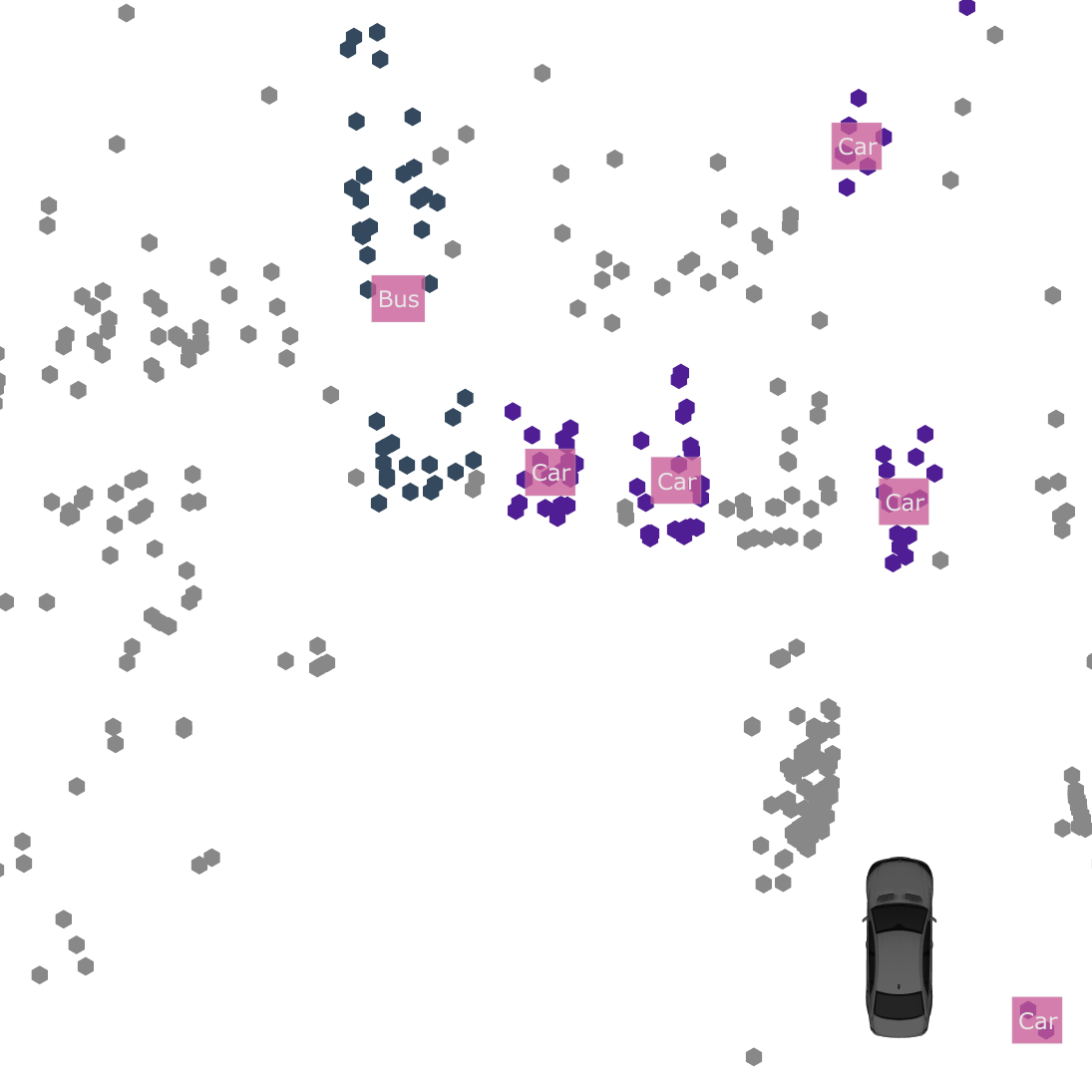}}
		\captionsetup{skip=1pt, font=footnotesize} \caption*{{Commercial Area}}
	\end{subfigure}

	\begin{subfigure}{.32\linewidth}
		\includegraphics[width=\textwidth]{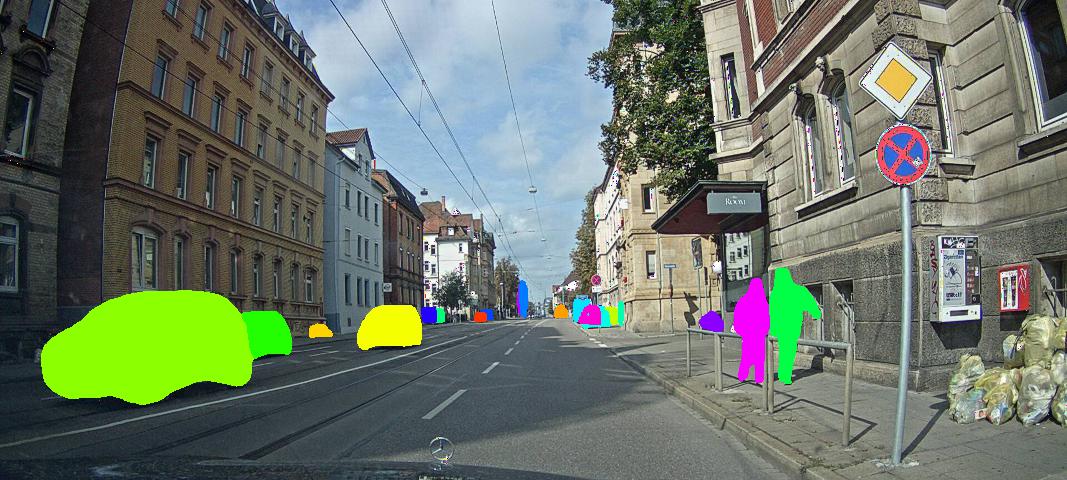}
	\end{subfigure}
	\begin{subfigure}{.32\linewidth}
		\includegraphics[width=\textwidth]{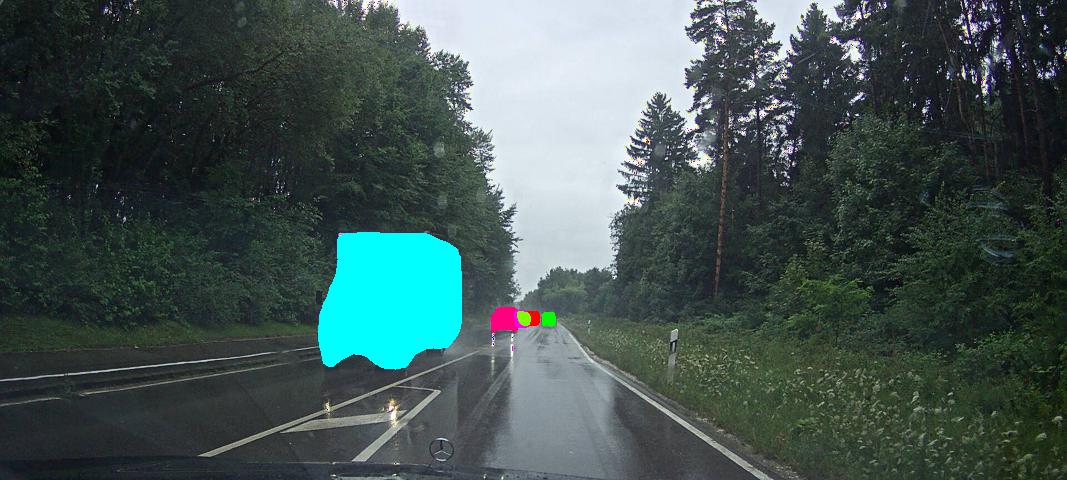}
	\end{subfigure}
	\begin{subfigure}{.32\linewidth}
		\includegraphics[width=\textwidth]{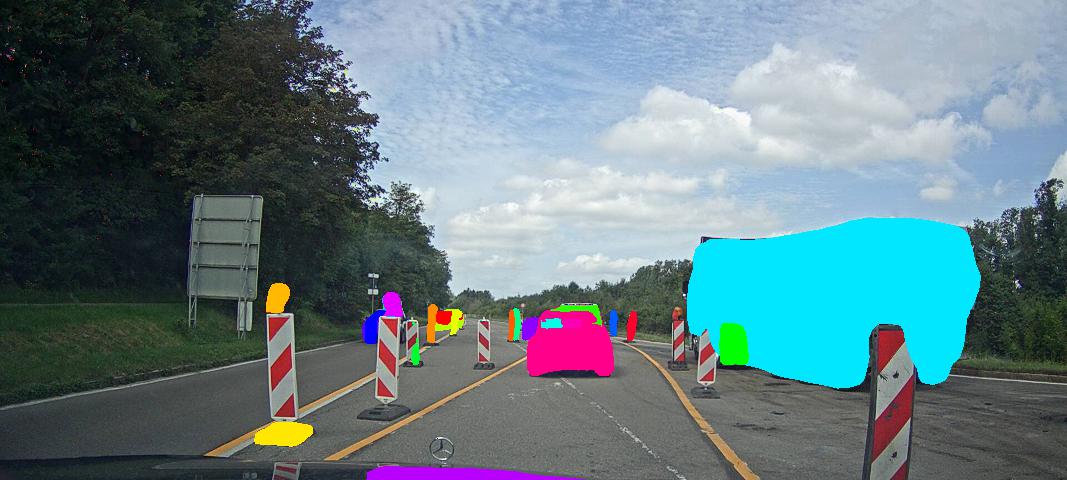}
	\end{subfigure}
	\vspace{.5em}
	
	\begin{subfigure}{.32\linewidth}
		\frame{\includegraphics[width=\textwidth]{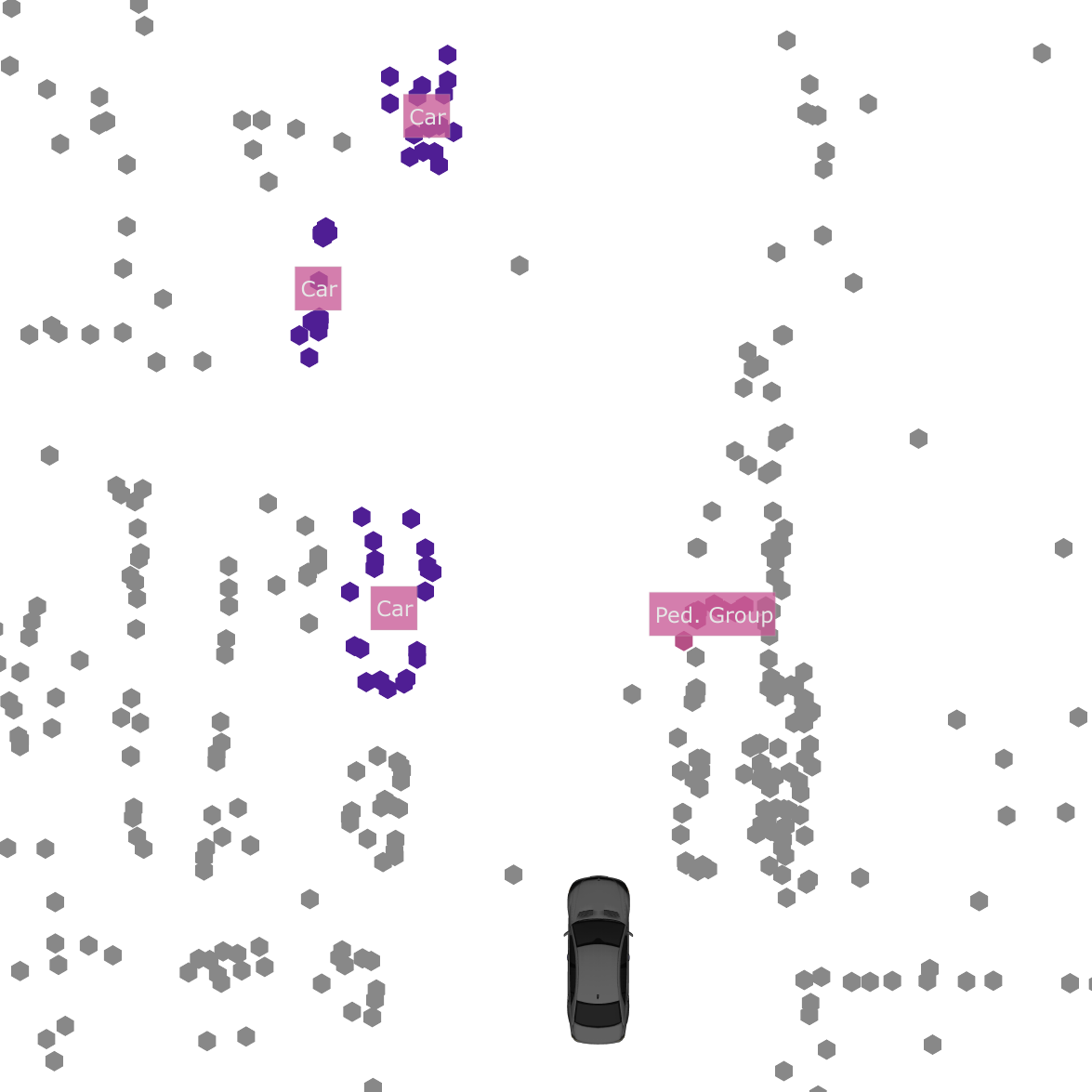}}
		\captionsetup{skip=1pt, font=footnotesize} \caption*{{Urban Area}}
	\end{subfigure}
	\begin{subfigure}{.32\linewidth}
		\frame{\includegraphics[width=\textwidth]{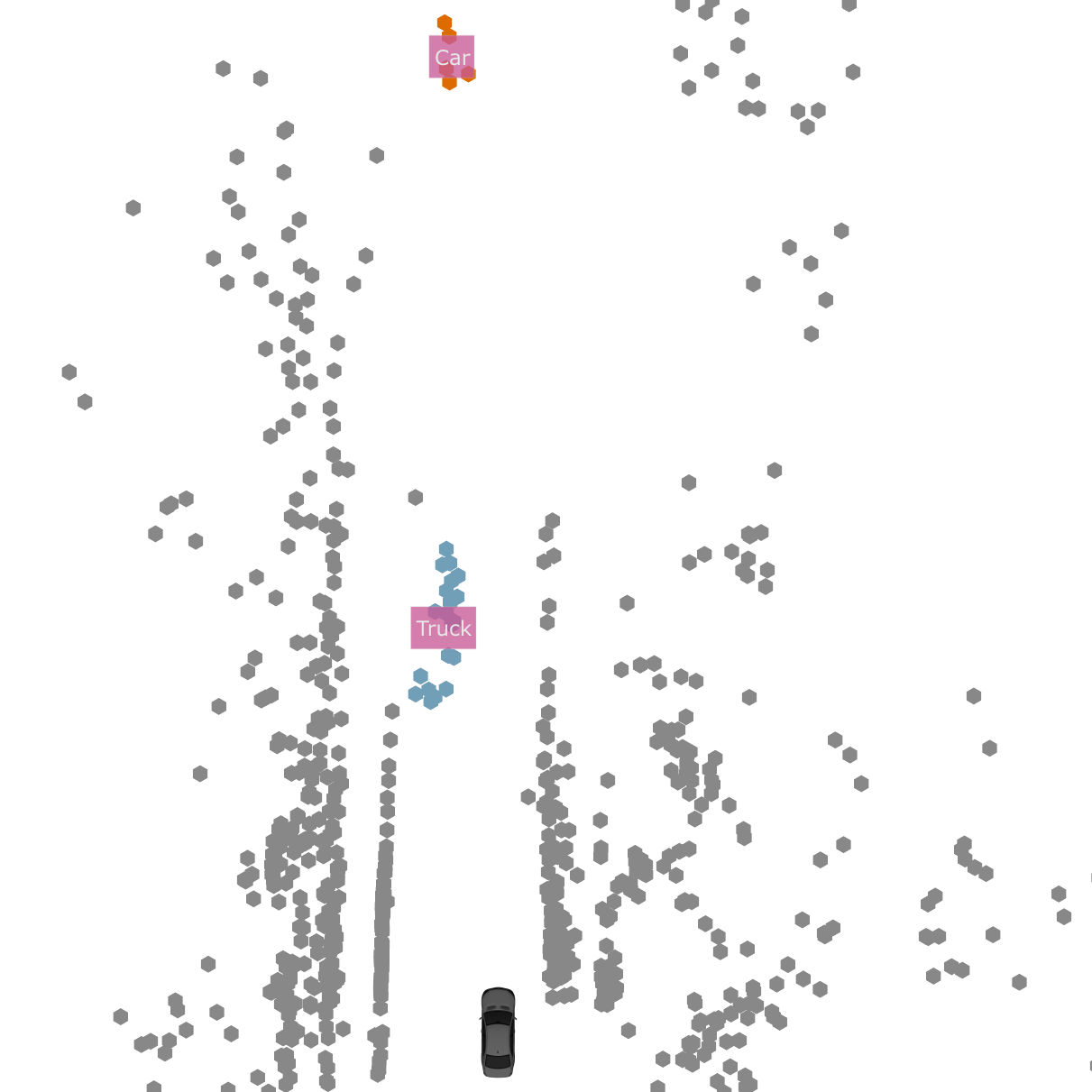}}
		\captionsetup{skip=1pt, font=footnotesize} \caption*{{Country Road}}
	\end{subfigure}
	\begin{subfigure}{.32\linewidth}
		\frame{\includegraphics[width=\textwidth]{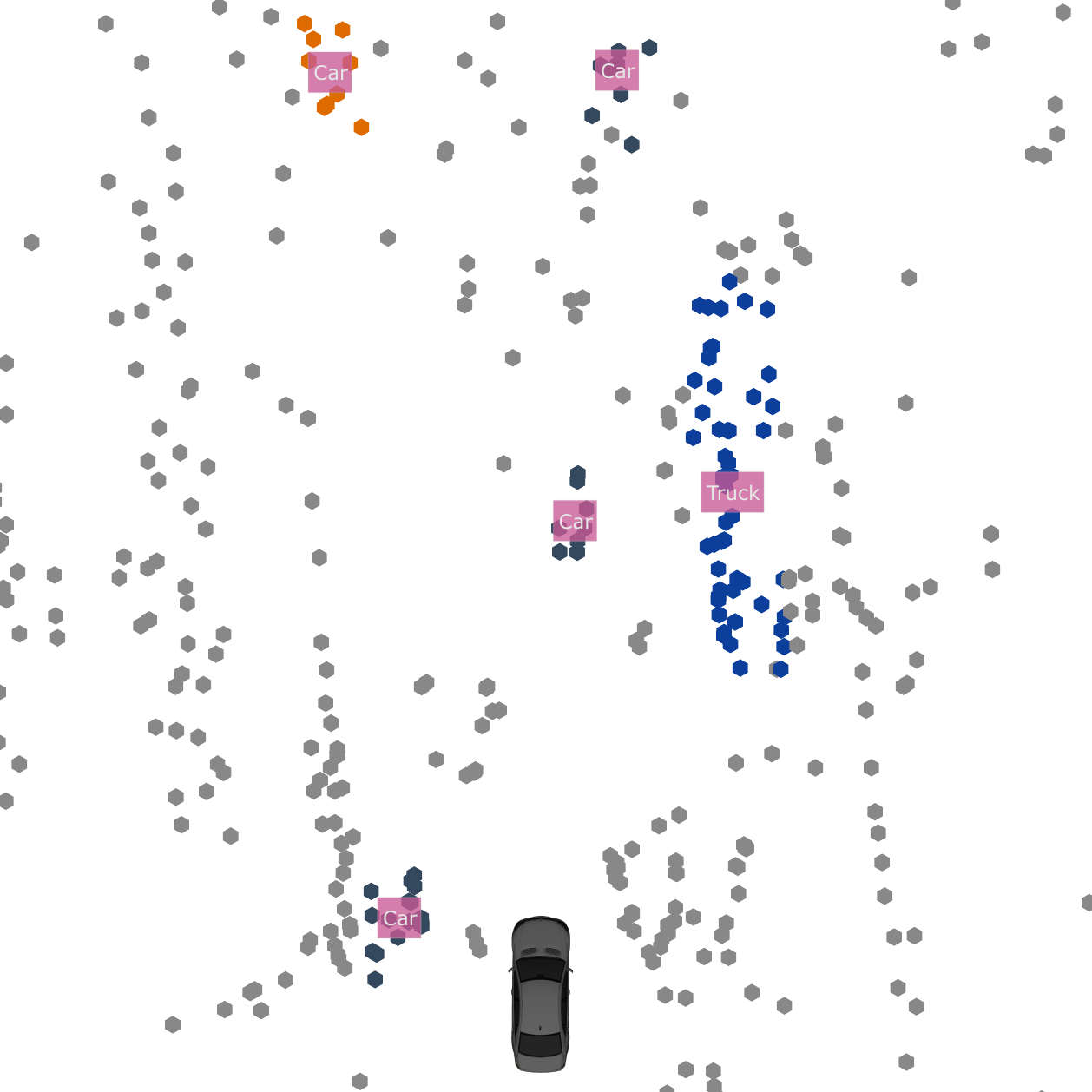}}
		\captionsetup{skip=1pt, font=footnotesize} \caption*{{Road Works}}
	\end{subfigure}
	\caption{This article introduces the first diverse large-scale data set for automotive radar point clouds. It comprises over 4 hours and \SI{100}{\km} of driving at a total of 158 different sequences with over 7500 manually annotated unique road users from 11 different object classes.}
	\label{fig:abstract}
\end{figure}

Despite that, algorithms for a semantic scene understanding based on radar data
are still in their infancy. Contrary, the large community in the field of
computer vision has driven machine learning (ML) in a relatively short time to
such a high level of maturity that it already made its way into several
automotive applications, from advanced driver assistance systems (ADAS) to fully
autonomous driving.
A lot of new developments come from the advances in ML-based image processing.
Driven by strong algorithms and large open data sets, image-based object detection has been completely revolutionized in the last decade.
In addition to camera sensors, also lidar data can frequently be found in well-known automotive data sets such as KITTI \cite{Geiger2012CVPR}.
Probably due to the extreme labeling effort and limited availability of adequate radar sensors, no publicly available automotive radar sets were available until recently.
Nevertheless, ML-based automotive radar perception has made vast progress in the last few years \cite{Dickmann2019}.
Radar has been used, e.g., for classification \cite{Schumann2017, Scheiner2019IV, Tilly2021}, tracking \cite{Tilly2020, Scheiner2019IRS, Pegoraro2020}, or object detection \cite{Schumann2019, Scheiner2020FUSION, Palffy2020}.
These advancements have led to a situation where the limited amount of available data is now a key challenge for radar-based ML algorithms \cite{Scheiner2020ATZ}.
As an effect, many proprietary data sets have been developed and utilized.
This has two main disadvantages for major progress in the field:
It takes a lot of effort to record and label a representative and sufficiently large data set.
Moreover, it is extremely difficult for researchers to compare their results to other methods.

By now, a few automotive radar data sets have been introduced, however, most are still lacking in size, diversity, or quality of the utilized radar sensors.
To this end, in this article the new large-scale \emph{RadarScenes} data set \cite{radar_scenes} for automotive radar point cloud perception tasks is introduced, see Fig.~\ref{fig:abstract}.
It consists of 11 object classes with 5 large main categories and a total of over 7000 road users which are manually labeled on \SI{100}{\km} of diverse street scenarios.
The data set and some additional tools are made publicly available\footnote{Online available: \url{www.radar-scenes.com}}.

\section{Related Work} \label{sec:relwork}
From other automotive perception domains, there exist several important data sets like KITTI \cite{Geiger2012CVPR}, Cityscapes \cite{Cordts2016Cityscapes}, or EuroCity Persons \cite{braun2019eurocity}.
These data sets comprise a large number of recordings but contain only video and sometimes lidar data.
In automotive radar, the list of publicly available data sets has recently been extended by a couple of new candidates.
However, most of them are very small, use a special radar sensor, or have other limitations.
Currently, the following open radar data sets are available:
nuScenes \cite{nuscenes2019}, Astyx \cite{astyxDataset}, Oxford Radar RobotCar \cite{RadarRobotCarDatasetICRA2020}, MulRan \cite{mulran2020}, RADIATE \cite{radiate2020}, CARRADA \cite{Carrada2020}, Zendar \cite{zendar2020}, NLOS-Radar \cite{Scheiner2020CVPR}, and the CRUW data set \cite{wang2021rodnet}.
Despite the increasing number of candidates, most of them are not applicable to the given problem for various reasons.

The Zendar \cite{zendar2020} and nuScenes \cite{nuscenes2019} data sets use real world recordings for a reasonable amount of different scenarios.
However, their radar sensors offer an extremely sparse output for other road users which seems not reasonable for purely radar-based evaluations. The popular nuScenes data set can hence not be used properly for radar-only classification algorithms.
In the Zendar data set, a synthetic aperture radar (SAR) processing approach is used to boost the resolution with multiple measurements from different ego-vehicle locations.
While the benefits of their approach are clearly showcased in the accompanying article, questions about the general robustness for autonomous driving are not yet answered. Additionally, SAR is a technique for the \emph{static} environment whereas this data set focuses on \emph{dynamic} objects.

In the Astyx data set \cite{astyxDataset}, a sensor with much higher native resolution is used.
However, similar to Zendar, it mainly includes annotated cars and the sizes of the bounding boxes suffer from biasing effects.
Furthermore, the number of scenarios is very low and non-consecutive measurements do not allow any algorithms that incorporate time information.
A similar radar system to the Astyx one is used in the NLOS-Radar data set \cite{Scheiner2020CVPR}, though only pedestrians and cyclists are included in very specific non-line-of-sight scenarios.

In contrast to all previous variants, the Oxford Radar RobotCar \cite{RadarRobotCarDatasetICRA2020}, MulRan \cite{mulran2020}, and RADIATE \cite{radiate2020} data sets follow a different approach.
By utilizing a rotating radar which is more commonly used on ships, they create radar images which are much denser and also much easier to comprehend.
As a consequence, their sensor installation uses much more space and requires to be mounted on vehicle roofs similar to rotating lidar systems.
An additional downside of these data sets is the absence of Doppler information.
As the capability to measure radial velocities is one of the greatest advantages
of a radar sensor over a series production lidar, the usability of this radar type for automotive applications is diminished.

Lastly, CARRADA \cite{Carrada2020} and CRUW \cite{wang2021rodnet} also use conventional automotive radar systems.
Unfortunately, their annotations are only available for radar spectra, i.e., 2D projections of the radar data cube.
The full point cloud information is not included and for the CRUW data set only a small percentage of all sequences contains label information.

The aim of the presented data set is to provide the community with a large-scale real world data set of automotive radar data point clouds.
It includes point cloud labels for a total of eleven different classes as well as an extensive amount of recordings from different days, weather, and traffic conditions.
A summary of the main aspects of all discussed radar data sets is given in Tab.~\ref{tab:datasets_summary}.
From this table it can be gathered that the introduced data set is the most diversified among the ones which use automotive radars with higher resolutions.
This diversity is expressed in terms of the amount of major classes, driving scenarios, and solvable tasks.
While the size of the data set suffices to train state-of-the-art neural networks, its main disadvantage is the absence of other sensor modalities except for the documentation camera and odometry data required for class annotation and processing of longer sequences.
Therefore, its usage is limited to radar perception tasks or for pre-training a multi-sensor fusion network.

\begin{table*}[t!]
	\renewcommand{\arraystretch}{1.6}
	\caption{Overview of publicly available radar data sets: The columns indicate some main evaluation criteria for the assessment of an automotive radar data set for machine learning purposes. Only road user classes are considered for the object category counts. The \textit{varying scenarios} column combines  factors such as weather conditions, traffic density, or road classes (highway, suburban, inner city) and the column \textit{sequential data} indicates if temporal coherent sequences are available.}
	\label{tab:datasets_summary}
	\centering
	\resizebox{0.98\linewidth}{!}{%
		\begin{tabular}{|l|P{1.1cm}|P{2.2cm}|P{1.8cm}|P{1.2cm}|P{1.2cm}|P{1.2cm}|P{1.2cm}|H{1.2cm}H{1.2cm}P{2cm}|P{1.2cm}|}
			\toprule
			Data Set & Size & Radar Type & Other Modalities & Sequential Data & Includes Doppler & Object Categories & Class Balancing & Main Object Categories & Additional Categories & Object Annotations & Varying Scenarios \\
			\midrule
			Oxford Radar RobotCar \cite{RadarRobotCarDatasetICRA2020} & Large & Mechanically Scanning & Stereo Camera, Lidar & \cmark & \xmark & 0 & N/A & 0 & 0 & \xmark & \textbf{+} \\
			MulRan \cite{mulran2020} & Middle & Mechanically Scanning & Lidar & \cmark & \xmark & 0 & N/A & 0 & 0 & \xmark & \textbf{+} \\
			RADIATE \cite{radiate2020} & Large & Mechanically Scanning & Camera, Lidar & \cmark & \xmark & 8 & \textbf{+} & & & 2D Boxes & \textbf{++} \\
			\midrule
			nuScenes \cite{nuscenes2019} & Large & Low Res. Automotive & Camera, Lidar & \cmark & \cmark & 23 & \textbf{+} &  &  & 3D Boxes & \textbf{++} \\
			Zendar \cite{zendar2020} & Large & Low Res. Automo. (High Res. SAR) & Camera, Lidar & \cmark & \cmark & 1 & N/A & 1 & 0 & 2D Boxes (\SI{2.5}{\percent} Manual Labels) & \textbf{++} \\
			\midrule
			Astyx \cite{astyxDataset} & Small & Next Gen. Automotive & Camera, Lidar & \xmark & \cmark & 7 & \textbf{-} & 1 & 6 & 3D Boxes & \textbf{-} \\
			NLOS-Radar \cite{Scheiner2020CVPR} & Small & Next Gen. Automotive & Lidar & \cmark & \cmark & 2 & \textbf{++} & 2 & 2 & Point-Wise & \textbf{-} \\
			
			CARRADA \cite{Carrada2020} & Small & Automotive & Camera & \cmark & \cmark & 3 & \textbf{++} & 3 & 0 & Spectral Annotations & \textbf{-} \\
			CRUW \cite{wang2021rodnet} & Large & Automotive & Stereo Camera & \cmark & \xmark & 3 & \textbf{+} & 3 & 0 & \makecell{Spectral Boxes\\(Only \SI{19}{\percent})}& \textbf{++} \\
			\textbf{RadarScenes (Ours)} & Large & Automotive & Docu Camera & \cmark & \cmark & 11 & \textbf{+} & 5 & 6 & Point-Wise & \textbf{++} \\
			\bottomrule
		\end{tabular} %
	}
\end{table*}

\section{Data Set} \label{sec:dset}
\begin{figure*}[t!] 
	\centering
	\input{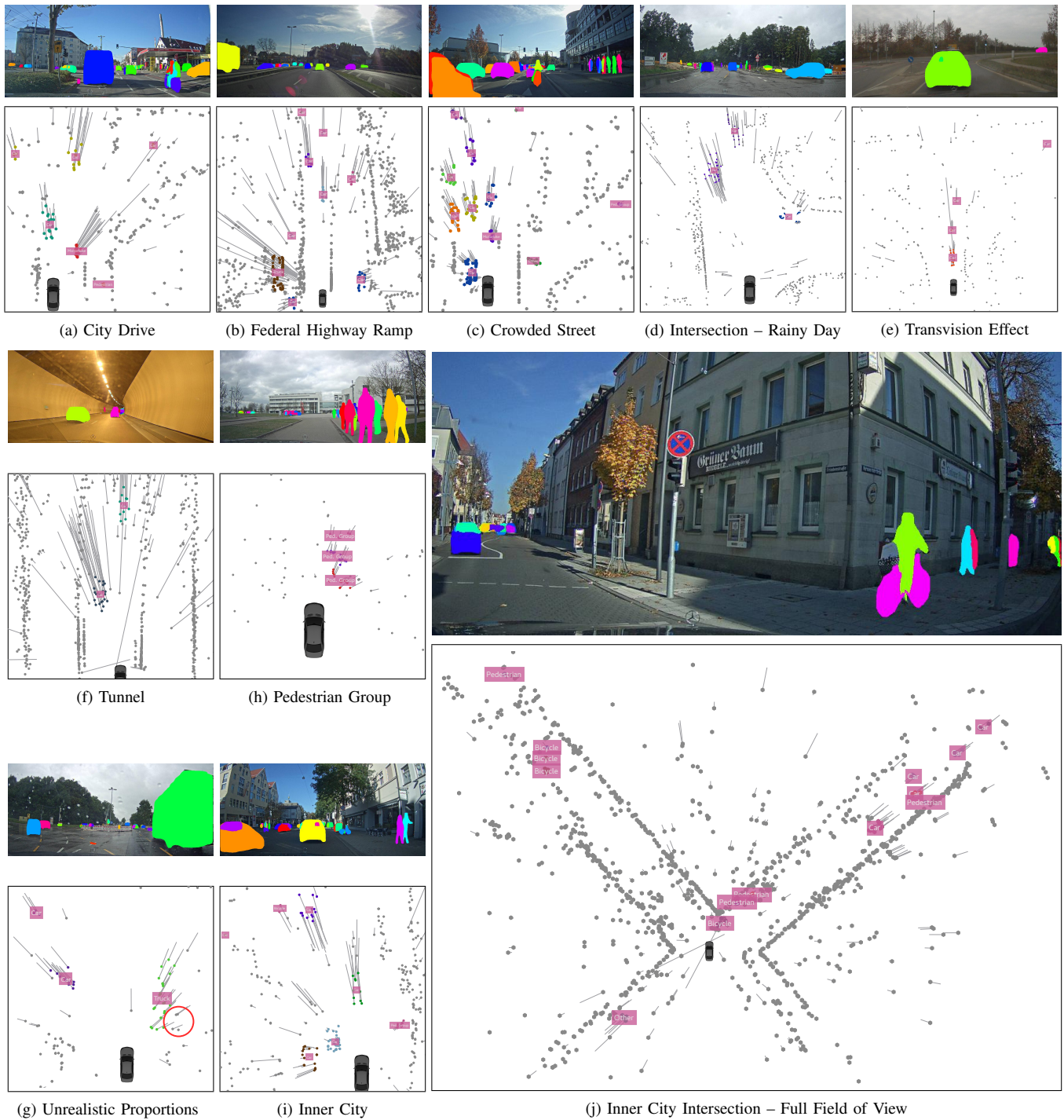}
	\caption{Radar bird's-eye view point cloud images and corresponding camera images. Orientation and magnitude of each point's estimated radial velocity after ego-velocity compensation are depicted by gray arrows. In the documentation camera images, road users are masked due to EU regulations (GDPR). Images allow for intensive zooming.}
	\label{fig:dset_examples}
\end{figure*}

In this section, the data set is introduced and important design decisions are discussed.
A general overview can be gathered from Fig.~\ref{fig:dset_examples}, where bird's eye view radar plots and corresponding documentation camera images are displayed.

\subsection{Measurement Setup}

\begin{figure}[ht!]
	\centering
	\input{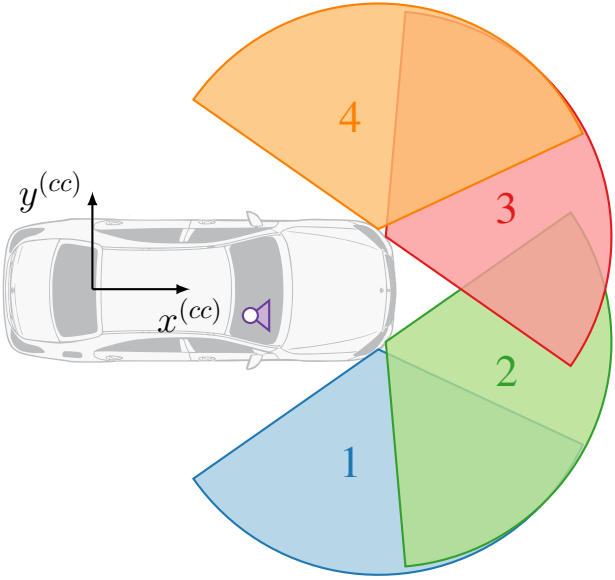}
	\caption{All sensor measurements are given in car coordinates ($cc$) with the
    origin located at the rear axle center. The documentary camera is mounted
    behind the windscreen (purple). The fields of view for each of the radar
    sensors are shown color-coded including their respective sensor ids.}
	\label{fig:measurement_setup}
\end{figure}

All measurements were recorded with an experimental vehicle equipped with four
\SI{77}{\giga\hertz} series production automotive radar sensors. While
supporting two operating modes, the data set consists only of measurements from
the near-range mode with a detection range of up to \SI{100}{\metre}. The
far-range mode has a very limited field of view and therefore is not considered
for the intended applications. Each sensor covers a $\pm \SI{60}{\degree}$ field
of view (FOV) in near-range mode, which is visualized in
Fig.~\ref{fig:measurement_setup} including their corresponding sensor ids. The
sensors with id 1 and 4 were mounted with $\pm \SI{85}{\degree}$ with respect to
the driving direction, 2 and 3 with $\pm \SI{25}{\degree}$.
In the sensor's data sheet the range and radial velocity accuracy are specified to be
$\Delta_r = \SI{0.15}{\metre}$ and $\Delta_v = \SI{0.1}{\kilo\metre\per\hour}$
respectively. For objects close to a boresight direction of the sensor, an
angular accuracy of $\Delta_{\phi}\left( \phi \approx \pm\SI{0}{\degree} \right) =
\SI{0.5}{\degree}$ is reached, whereas at the angular FOV borders it decreases
to $\Delta_{\phi}\left( \phi \approx \pm\SI{60}{\degree} \right) =
\SI{2}{\degree}$. Each radar sensor independently determines the measurement
timing on the basis of internal parameters causing a varying cycle time. On
average, the cycle time is \SI{60}{\milli\second} ($\approx \SI{17}{\hertz}$).

Furthermore, the vehicle's motion state (i.e. position, orientation, velocity,
and yaw rate) is recorded. Based on that, the radar measurement data can be
compensated for ego-motion and transformed into a global coordinate system.

For documentation purposes, a camera provides optical images of the recorded
scenarios. It is mounted in the passenger cabin behind the windshield (cf.
Fig.~\ref{fig:measurement_setup}). In order to comply with all rules of the
general data protection regulation (GDPR), other road users are anonymized by
repainting. Well-known instance segmentation neural networks were used to automate the
procedure, followed by a manual correction step.  Correction was done with the aim to remove false negatives, i.e., non-anonymized objects. False positive camera annotations may still occur.
Examples are given in Fig.~\ref{fig:dset_examples}.

\subsection{Labeling}
Data annotation for automotive radar point clouds is an extremely tedious task.
As can be gathered from Fig.~\ref{fig:dset_examples}, it requires human experts with a lot of experience to accurately identify all radar points which correspond to the objects of interest.
In other data sets, the labeling step is often simplified.
For example, CARRADA and CRUW use a semi-automatic annotation approach transforming camera label predictions to the radar domain.
In the NLOS-Radar data set, instructed test subjects carry a GNSS-based reference system from \cite{Scheiner2019IRS} for automatic label generation.
Moreover, lidar data is a common source for generating automated labels.
To ensure high quality annotations and not have to rely on instructed road users, the introduced data set uses only manually labeled data created by human experts.

In automotive radar, the perception of static objects and moving road users is often separated into two independent tasks.
The reason for this is that for stationary objects longer time frames of sparse radar measurement can be accumulated than for volatile dynamic targets.
This increases the data density, which facilitates the perception tasks for static objects.
The segregation is enabled by the Doppler velocities, which allow to filter out moving objects.
The introduced data set is focused on moving road users. Therefore, eleven different object classes are labeled: \texttt{car}, \texttt{large vehicle}, \texttt{truck}, \texttt{bus}, \texttt{train}, \texttt{bicycle}, \texttt{motorized two-wheeler}, \texttt{pedestrian}, \texttt{pedestrian group}, \texttt{animal}, and \texttt{other}.
The \texttt{pedestrian group} label is assigned where no certain separation of individual pedestrians can be made, e.g., in Fig. \ref{fig:ped_grp}.
The \texttt{other} class consists of various other road users that do not fit well in any of these categories, e.g., skaters or moving pushcarts.
In addition to the eleven object classes, a \texttt{static} label is assigned to all remaining points.
All labels were chosen in a way to minimize disagreements between different experts' assessment, e.g., a large pickup van may have been characterized as \texttt{car} or \texttt{truck} if no \texttt{large vehicle} label had been present.
In general, the exact choice of label categories always poses a problem.
In \cite{Scheiner2019EuMW}, it is analyzed how even different types of bicycles can affect the results of classifiers.
Instead of finding more and more fine-grained sub-classes, the impact of this problem can also be alleviated by introducing more variety within the individual classes.

To this end, in addition to the regular 11 classes, a mapping function is provided along with the data set tools which project the majority of classes to a more coarse variant containing only: \texttt{car}, \texttt{large vehicle}, \texttt{two-wheeler}, \texttt{pedestrian}, and \texttt{pedestrian group}.
This alternative class definition can be utilized for tasks, where a more balanced label distribution is required to achieve good performance, e.g., for classification or object detection tasks, e.g., \cite{Schumann2017, Schumann2019}.
For this mapping, the \texttt{animal} and the \texttt{other} class are not reassigned.
Instead they may be used, e.g., for a hidden class detection task, cf. \cite{Scheiner2019IV}.

Two labels are assigned to each individual detection of a dynamic object: a \texttt{label id} and a \texttt{track id}. The \texttt{label id} is an integer describing the semantic class the respective object belongs to. On the other hand, the \texttt{track id} identifies single real world objects uniquely over the whole recording time. That is, all detections of the same object have the same \texttt{track id}. This allows to develop algorithms which take as input not only data from one single radar measurement but from a sequence of measurement cycles.
If an object was occluded for more than \SI{500}{ms} or if the movement stopped for this period of time (e.g. if a car stopped at a red light), then a new \texttt{track id} is assigned as soon as detections of the moving object are measured again.

The labelers were instructed to label the objects in such a way that realistic and consistent object proportions are maintained. Since automotive radar sensors have a rather low accuracy in the azimuth direction, especially in greater distances the contour of objects is not captured accurately.
For example, deviations in the azimuth direction can cause that an object in front of the ego-vehicle appears much wider than it actually is.
In this case, detections with a non-zero Doppler velocity are reported at positions which are unrealistically far away from the true object's position. These detections are then not added to the object.
One example for this situation is depicted in Fig. \ref{fig:labeling_size}, where only those detections matching the \texttt{truck}'s true width were annotated. The detections lying in the encircled area were deliberately excluded, even though they have a non-zero Doppler velocity.
An object's true dimensions were estimated by the labelers over the whole time during which detections from the object were measured.
This approach causes that at first sight the labeling results seem wrong, since detections with non-zero Doppler close to a moving object are not marked as members of this object.
However, this visual impression is in most cases misleading and great care was taken that object dimensions are consistent over time.
This choice has the great upside that realistic object extents can be estimated by object detection algorithms.

Mirror effects, ambiguities in each measurement dimension as well as false-positive detections from large sidelobes can result in detections with non-zero Doppler which do not belong to any moving object. These detections have the default label \texttt{static}. Hence, simply thresholding all detections in the Doppler dimension does not result in the set of detections that belong to moving road users.

Each detection in the data set can be identified by a universally unique identifier (UUID). This information can be used for debugging purposes as well as for assigning predicted labels to detections so that a link to the original measurement is created and visualization is eased.

\subsection{Statistics}
The data set contains a total of $118.9$ million radar points measured on $\SI{100.1}{\kilo\metre}$ of urban and rural routes.
The data were collected from various scenarios with varying durations from $\SI{13}{\second}$ to $\SI{4}{\minute}$ accumulating to a total of $\SI{4.3}{\hour}$. 
In comparison, KITTI \cite{Geiger2012CVPR}  contains only about $\SI{1.5}{\hour}$ of data.
The overall size of the data set competes with the nuScenes \cite{nuscenes2019} data set, which is one of the largest publicly available driving data sets with radar measurements  and which contains about $1.4$M 3D bounding box annotations in a total measurement time of $\SI{5.5}{\hour}$.
But as stated before, the radar point cloud provided in the nuScenes data set is very sparse due to a different data interface and hence stand-alone radar perception algorithms can only be developed to some extent.
A more fine-grained overview of the statistics of this data set is given in Tab.~\ref{tab:dataset_stats}.
In this table, detections originating from the static environment are left out since this group constitutes the fallback class and comprises more than $\SI{90}{\percent}$ of all measured points. 
Of the annotated classes, the majority class is the \texttt{car} class followed by \texttt{pedestrians} and \texttt{pedestrian groups}. 
Those three classes account for more than $\SI{80}{\percent}$ of the unique objects resulting in an imbalance between classes in the data set. 
The imbalance problem is common among driving data sets and can for example also be found in the nuScenes \cite{nuscenes2019} or KITTI \cite{Geiger2012CVPR} data sets.
However, in other data sets even more extreme imbalances can be found, see Tab. \ref{tab:datasets_summary}.
The handling of this problem is one major aspect for robust detection and classification in the area of automated driving.
In Fig.~\ref{fig:data_stats}, additional information about the distribution of object annotations is depicted with respect to the entire data set, the sensor scan, and the object distances to the ego-vehicle.

\begin{table*}[t!]
	\caption{Data set statistics per class before (upper) and after (lower) class mapping.}
	\label{tab:dataset_stats}
	\centering
	\resizebox{0.995\linewidth}{!}{%
	\begin{tabular}{lrrrrrrrrrrrr}
\toprule
{} & \thead{CAR} & \thead{LARGE VEHICLE} & \thead{TRUCK} & \thead{BUS} & \thead{TRAIN} & \thead{BICYCLE} & \thead{MOTORBIKE} & \thead{PEDESTRIAN} & \thead{PEDESTRIAN GROUP} & \thead{ANIMAL} & \thead{OTHER} & \thead{$\sum$} \\
\midrule
Annotations $(10^3)$ &  $1677$  $(\SI{42.01}{\percent})$ &          $43$  $(\SI{1.08}{\percent})$ &  $545$ $(\SI{13.64}{\percent})$ &  $211$  $(\SI{5.28}{\percent})$ &    $6$  $(\SI{0.14}{\percent})$ &   $199$  $(\SI{4.97}{\percent})$ &                   $9$  $(\SI{0.23}{\percent})$ &      $361$  $(\SI{9.04}{\percent})$  &            $864$  $(\SI{21.63}{\percent})$ &  $0.5$  $(\SI{0.01}{\percent})$ &   $78$  $(\SI{1.96}{\percent})$ &  $3992$ \\
Unique Objects       &  $3439$  $(\SI{45.76}{\percent})$ &          $49$  $(\SI{0.65}{\percent})$ &  $346$ $(\SI{4.60}{\percent})$  &   $53$  $(\SI{0.71}{\percent})$ &    $2$  $(\SI{0.03}{\percent})$ &   $268$  $(\SI{3.57}{\percent})$ &                  $40$  $(\SI{0.53}{\percent})$ &     $1529$  $(\SI{20.34}{\percent})$ &           $1124$  $(\SI{14.95}{\percent})$ &    $4$  $(\SI{0.05}{\percent})$ &  $662$  $(\SI{8.81}{\percent})$ &  $7516$ \\
Time (s)             & $23231$  $(\SI{38.56}{\percent})$ &         $418$  $(\SI{0.69}{\percent})$ & $3133$ $(\SI{5.20}{\percent})$  &  $829$  $(\SI{1.38}{\percent})$ &   $14$  $(\SI{0.02}{\percent})$ &  $3261$  $(\SI{5.41}{\percent})$ &                 $198$  $(\SI{0.33}{\percent})$ &    $11627$  $(\SI{19.30}{\percent})$ &          $14890$  $(\SI{24.72}{\percent})$ &   $19$  $(\SI{0.03}{\percent})$ & $2621$  $(\SI{4.35}{\percent})$ & $60239$ \\
\midrule
& \multicolumn{1}{c}{$\underbrace{\hspace{1.5cm}}$} & \multicolumn{4}{c}{$\underbrace{\hspace{7.7cm}}$} &  \multicolumn{2}{c}{$\underbrace{\hspace{4cm}}$} & \multicolumn{1}{c}{$\underbrace{\hspace{1.5cm}}$} & \multicolumn{1}{c}{$\underbrace{\hspace{2.2cm}}$} & & & \\
{} & \thead{CAR} &  \multicolumn{4}{c}{\thead{LARGE VEHICLE}}  & \multicolumn{2}{c}{\thead{TWO-WHEELER}} & \thead{PEDESTRIAN} & \thead{PEDESTRIAN GROUP} & & & \thead{$\sum$} \\
\midrule
Annotations $(10^3)$ &  $1677$  $(\SI{42.84}{\percent})$ &   \multicolumn{4}{c}{$805$ $(\SI{20.56}{\percent})$} &   \multicolumn{2}{c}{$208$  $(\SI{5.31}{\percent})$}  &      $361$  $(\SI{9.22}{\percent})$  &            $864$  $(\SI{22.07}{\percent})$ & \multicolumn{1}{c}{-} & \multicolumn{1}{c}{-} &  $3915$ \\
Unique Objects       &  $3439$  $(\SI{50.20}{\percent})$ & \multicolumn{4}{c}{$450$ $(\SI{6.57}{\percent})$} & \multicolumn{2}{c}{$308$  $(\SI{4.50}{\percent})$} &     $1529$  $(\SI{22.32}{\percent})$ &           $1124$  $(\SI{16.41}{\percent})$ & \multicolumn{1}{c}{-} & \multicolumn{1}{c}{-} &  $6850$ \\
Time (s)             & $23231$  $(\SI{40.33}{\percent})$ & \multicolumn{4}{c}{$4394$ $(\SI{7.63}{\percent})$} & \multicolumn{2}{c}{$3459$  $(\SI{6.00}{\percent})$} &    $11627$  $(\SI{20.19}{\percent})$ &          $14890$  $(\SI{25.85}{\percent})$ & \multicolumn{1}{c}{-} & \multicolumn{1}{c}{-} & $57601$ \\
\bottomrule
\end{tabular}
	}
\end{table*}
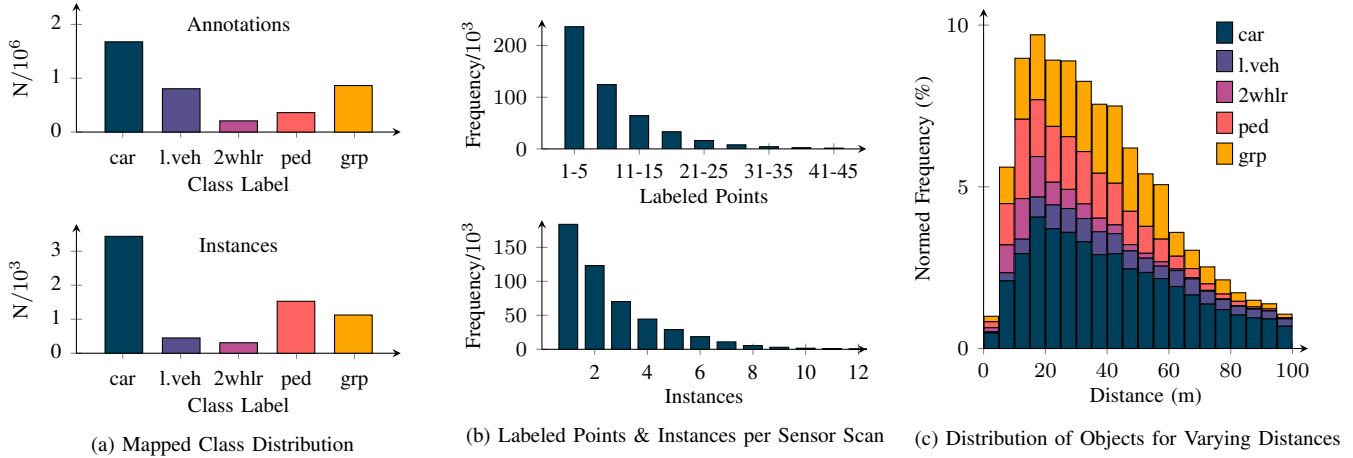
\begin{figure*}[t!]
	\begin{subfigure}{.325\textwidth}
\begin{subfigure}{\textwidth}
\begin{tikzpicture}
\begin{axis}[   title={Annotations},
    			title style={font=\footnotesize,yshift=-4.5ex},
                ybar,
                width=\textwidth,
                bar width=0.5cm,
                height=3.3cm,
                ymin=0,
                ymax=2400,
                xmin=0.2,
                xmax=5.8,
				ytick={0,1000,2000},
				yticklabels={0,1,2},
				xtick={1,2,3,4,5},
				xticklabels from table={figures/tikz_data_map_cls.txt}{cls},
                axis lines=left,
                tick pos=left,
                xlabel={Class Label},
                ylabel={N$/10^{6}$}, 
                ylabel near ticks,
                tick label style={font=\footnotesize}, 
                %x tick label style={yshift={-mod(\ticknum,2)*1em}},
                typeset ticklabels with strut,
                label style={font=\footnotesize},    
             	x label style={font=\footnotesize,yshift=.4ex},   
        ]
        \addplot [fill=barblue2] coordinates {(2.5,1677)}; 
        \addplot [fill=barblue] coordinates {(2.75,805)}; 
        \addplot [fill=barpink] coordinates {(3,208)}; 
        \addplot [fill=barred] coordinates {(3.25,361)}; 
        \addplot [fill=barorange] coordinates {(3.5,864)};
        %\addplot [fill=barblue] table[x=cls_id,y=amount,col sep=space]{figures/tikz_data_cls.txt};
\end{axis}

\end{tikzpicture}
%\captionsetup{skip=5pt, font=footnotesize} \caption{{Class Distribution}}
\end{subfigure}
\vspace{.9em}

\begin{subfigure}{\textwidth}
	\begin{tikzpicture}
	\begin{axis}[
	title={Instances},
 	title style={font=\footnotesize,yshift=-4.5ex},
	ybar,
	width=\textwidth,
	bar width=0.5cm,
	height=3.3cm,
	ymin=0,
	ymax=3800,
	xmin=0.2,
	xmax=5.8,
	ytick={0,1000,2000,3000},
	yticklabels={0,1,2,3},
	xtick={1,2,3,4,5},
	xticklabels from table={figures/tikz_data_map_cls.txt}{cls},
	axis lines=left,
	tick pos=left,
    xlabel={Class Label},
	ylabel={N$/10^{3}$}, 
	ylabel near ticks,
	tick label style={font=\footnotesize}, 
	typeset ticklabels with strut,
	label style={font=\footnotesize},     
	x label style={font=\footnotesize,yshift=.4ex},  
	]
    \addplot [fill=barblue2] coordinates {(2.5,3439)};	
    \addplot [fill=barblue] coordinates {(2.75,450)}; 
    \addplot [fill=barpink] coordinates {(3,308)}; 
    \addplot [fill=barred] coordinates {(3.25,1529)}; 
    \addplot [fill=barorange] coordinates {(3.5,1124)};
	\end{axis}
	
	\end{tikzpicture}
	\captionsetup{skip=5pt, font=footnotesize} \caption{{Mapped Class Distribution}}
	\label{fig:class_dist}
\end{subfigure}
\end{subfigure}
\begin{subfigure}{.325\textwidth}
	\begin{subfigure}{\textwidth}
\begin{tikzpicture}
\begin{axis}[
                ybar,
                width=\textwidth,
                bar width=0.25cm,
                height=3.3cm,
                ymin=0,
                ymax=250,
                xmin=0,
                xmax=50,
                axis lines=left,
                tick pos=left,
                xlabel={Labeled Points},
                ylabel={Frequency/$10^3$},
                ylabel near ticks,
                tick label style={font=\footnotesize}, 
				x label style={font=\footnotesize,yshift=.8ex},  
				y label style={font=\footnotesize,yshift=-.8ex},  
				xtick={5,15,25,35, 45},
				xticklabels={1-5, 11-15, 21-25, 31-35, 41-45},
        ]
        \addplot [fill=barblue2] table[x=n,y=pts,col sep=space]{figures/tikz_data_pts_per_scan.txt};
\end{axis}

\end{tikzpicture}
\end{subfigure}

\begin{subfigure}{\textwidth}
\begin{tikzpicture}
\begin{axis}[
				ybar,
				width=\textwidth,
				bar width=0.25cm,
				height=3.3cm,
				ymin=0,
				ymax=190,
				xmin=0,
				xmax=12.3,
				axis lines=left,
				tick pos=left,
				xlabel={Instances},
				ylabel={Frequency/$10^3$},
				ylabel near ticks,
				tick label style={font=\footnotesize}, 
				x label style={font=\footnotesize,yshift=.8ex},  
				y label style={font=\footnotesize,yshift=-.8ex},  
				xtick={2,4,6,8, 10, 12},
				xticklabels={2, 4, 6, 8, 10, 12},
		]
		\addplot [fill=barblue2] table[x=n,y=inst,col sep=space]{figures/tikz_data_inst_per_scan.txt};
\end{axis}

\end{tikzpicture}
\captionsetup{skip=5pt, font=footnotesize} \caption{{Labeled Points \& Instances per Sensor Scan}}
\label{fig:frame_dist}
\end{subfigure}
\end{subfigure}
\begin{subfigure}{.325\textwidth}
\begin{tikzpicture}

\begin{axis}[
				width=\textwidth, % 6cm,
				height=6.1cm, % 5.1cm,
				axis lines=left,
				ybar stacked,   % Stacked vertical bars
				bar width=5.5pt,
				tick pos=left,
				ymin=0,         % Start y axis at 0
				ymax=10.5,
				xmin=0,
				xmax=105,
				legend columns=4, 
				legend style={%/tikz/column 2/.style={column sep=5pt,},
					fill=none,
					draw=none,
					font=\footnotesize,
				},
				legend columns=1,
				legend cell align={left},
				%legend pos=north east,
				ylabel={Normed Frequency (\%)}, 
				xlabel={Distance  (m)}, 	
				xtick=data,
				scaled ticks=false,
				xtick={0,20,40,60,80,100},
				ylabel near ticks,
				tick label style={font=\footnotesize}, 
				x label style={font=\footnotesize,yshift=.8ex},  
				y label style={font=\footnotesize,yshift=-.5ex},  
		]
\addplot [fill=barblue2] table [y=car_norm,x=bins,col sep=space] {figures/hist20.txt}; 
\addplot [fill=barblue] table [y=trk_norm,x=bins,col sep=space] {figures/hist20.txt}; 
\addplot [fill=barpink] table [y=bik_norm,x=bins,col sep=space] {figures/hist20.txt}; 
\addplot [fill=barred] table [y=ped_norm,x=bins,col sep=space] {figures/hist20.txt}; 
\addplot [fill=barorange] table [y=grp_norm,x=bins,col sep=space] {figures/hist20.txt};

\legend{car,l.veh,2whlr,ped,grp};

\end{axis}
\end{tikzpicture}
\captionsetup{skip=5pt, font=footnotesize} \caption{{Distribution of Objects for Varying Distances}}
\label{fig:distance_dist}
\end{subfigure}
	\caption{Data set statistics for the five main object classes: (\subref{fig:class_dist}) indicates the amount of annotated points and individual objects for each class. Fig. (\subref{fig:frame_dist}) shows how points and instances are distributed over the sensors scans. In (\subref{fig:distance_dist}), the normed distributions of all five classes are displayed for ranges up to \SI{100}{\meter}.}
	\label{fig:data_stats}
\end{figure*}

\subsection{Data Set Structure}
For each sequence in the data set, radar and odometry data are contained in a single Hierarchical Data Format file (hdf5), in which both modalities are entered as separate tables.
Each row in the odometry table provides the global coordinates and orientation of the ego-vehicle for one timestamp, which is given in the first column.
In the radar data table, each row describes one radar detection with timestamp, position in the global, ego-vehicle and sensor coordinate system, 
ego-motion-compensated radial velocity and radar cross-section (RCS) as well as the annotation information given as unique \texttt{track id} and class \texttt{label id}.
By providing both, the local and global positions of the detection, coordinate
transformations do not have to be performed and detections could for example be accumulated directly in a map in the global coordinate frame.
The reference camera images are saved as JPG files in the camera folder for each sequence with the recording timestamp as their base name. 
The additional \textit{scenes.json} file in a JSON format maps the radar measurement timestamps to the reference camera image and odometry table row index and provides the timestamp (used as key) of the next measured frame, allowing for easy iteration through
the sequence.

\section{Evaluation} \label{sec:eval}
\emph{RadarScenes} allows to build and verify various types of applications.
Among others, the data set may be used for the development of algorithms in the fields object detection, object formation, tracking, classification, semantic (instance) segmentation and grid mapping.
With the provided annotations, the performance of these algorithms can be directly assessed.
The choice of an evaluation metric entirely depends on the problem that is to be solved.
In order to provide a common base, we suggest a couple of important evaluation measures which are supposed to ease the comparability of independent studies.
For all of the following three tasks, additionally a point-wise macro-averaged F1 score may be reported to allow for a comparison of the different approaches. 

\subsection*{Object Detection and Instance Segmentation}
Object detection and instance segmentation are two of the fastest growing fields.
Moreover, the quantification of test results requires several tweaks that a can be utilized for other applications, too.
Thus, these two disciplines are derived first.

The sparsity of automotive radar point clouds usually requires an accumulation of multiple sensor scans.
This is beneficial for signal processing, however, it makes the comparison of new approaches difficult, especially when using different accumulation windows or update rates.
Moreover, the radar point clouds in this data set are only labeled point-wise,
i.e., no bounding box with real object dimensions is represented, as these often
cannot be estimated accurately without the excessive support from other depth sensors.
In order to overcome these issues and to provide a unified evaluation benchmark for various algorithmic approaches, the following evaluation scheme is proposed:
First, we do not discriminate between object detection and instance segmentation.
Object detection algorithms can be applied, but the evaluation will treat object detectors as if they were instance segmentation models.
Therefore, results are evaluated on a point level utilizing the intersection of the predicted instances with the ground truth instances and their class labels at a given time step.
Second, the evaluation is carried out at every available sensor cycle for any object present in that sensor scan.
For model design, arbitrary accumulation schemes may be chosen, but the results should be inferred at the exact time step when the radar point is measured and every radar point is only evaluated once. That is, if a single detection is included in multiple prediction cycles because accumulated data is used, then only the \emph{first} prediction of this detection shall be considered.
This way, it can be ensured, that no future data finds its way into the evaluation of a specific object.
Moreover, any accumulation or recurrent feedback structure can be broken down to this evaluation scheme on a common time base.
This does however not imply that it makes sense to build a model that needs, e.g., \SI{30}{\second} of accumulated data to function properly even if this may improve the results on specific scenarios.

Since using all 11 object categories would lead to a quite imbalanced training and validation set, we recommend to use the mapped labels, i.e., the five main object classes.
Classes that are not included in the mapped version, i.e. \texttt{animal} and \texttt{other}, can be excluded during training and evaluation.

The most prominent metric in this area is mean Average Precision (mAP) \cite{pascal-voc-2008} evaluated at an intersection over union (IoU) of typically $0.5$, see \cite{Palffy2020} for a detailed explanation on how the IoU concept can be utilized for radar point clouds.
Another common metric, which is mainly applied for small objects such as vulnerable road users, is the log-average miss rate (LAMR) \cite{Dollar2012}.
It is used to shift the detector's focus more on the recall and not the precision.
Another commonly used metric, especially in the radar domain, is the class-averaged F1 score.
Similar to mAP, averaging occurs after calculating the individual class F1 scores, i.e., macro-averaging.
To increase the comparability, we encourage to report several metrics for experiments on new models, but at least the $\text{mAP}_{50}$ measure.

Moreover, we want to highlight the fact that radar data in general -- and therefore also in this data set -- often contains data points which appear to be part of an object, but are actually caused by a different reflector, e.g., ground or multi-path detections.
Box predictors such as commonly used in image-based object detection often have difficulties in reaching high IoU values with ground truth objects, even for seemingly perfect predictions.
Therefore, it may be beneficial to include additional evaluations for lower IoU threshold, e.g., at $0.3$.

\subsection*{Classification}
The main difference between instance segmentation and classification is that not a whole scene is used as input but rather only object clusters are used, which might result from either a clustering algorithm or the ground truth annotations.
The same frame accumulation advises as for object detection apply.
Scores are reported as instance-based averaged F1 scores based on all five main classes and one rejection class.
The rejection class (\texttt{clutter} or even \texttt{static}) shall be used for those clusters that were erroneously created by a clustering algorithm and have no overlap with any ground truth object. The classifier's task is then to classify these clusters as \texttt{clutter}.
In addition, mAP on the object classes can be reported in a ceiling analysis for object detectors.
For hidden class detection on the \texttt{animal} and \texttt{other} class, regular precision and recall values should be reported alongside the F1 scores of the main classes.

\subsection*{Semantic Segmentation}
For semantic segmentation the same frame accumulation advises as for object detection apply.
Scores are reported as point-wise macro-averaged F1 scores based on all five main classes and one \texttt{static} background class.
Additional class-wise scores help to interpret the results.

\section{Conclusion} \label{sec:conclusion}
In this article, a new automotive radar data set is introduced.
Due to its large size, the realistic scenario selection as well as the point-wise annotation of both semantic \texttt{class id} and \texttt{track id} for moving road users, the data set can be directly used for the development of various (machine learning-based) algorithms.
The focus of this data set lies entirely on radar data.
Details on the data distribution and the annotation process are given.
Images from a documentary camera are provided to allow for an easier understanding of the scenes.
In addition to the data set, we propose a detailed evaluation scheme, allowing to compare different techniques even when using different accumulation schemes.
We cordially invite other researches to publish the algorithms and scores they achieve on this data set for tasks like object detection, clustering, classification or semantic (instance) segmentation.
Our hope is that this data set inspires other researchers to work in this field.

% ###################################
\bibliographystyle{IEEEtran}
\bibliography{mybibfile}
\end{document}